\documentclass[letterpaper, 10 pt, conference]{ieeeconf}  

\IEEEoverridecommandlockouts                              
                                                          
\usepackage{xcolor} 
\usepackage{graphicx} 
\graphicspath{{./figures/}}
\usepackage{amsmath} 
\usepackage{amssymb}  
\usepackage{algpseudocode}
\usepackage{algorithm} 
\usepackage{wrapfig}
\usepackage{hyperref}
\usepackage{cite}
\usepackage{eso-pic}
\newcommand{\algo}{FAMLE} 
\newcommand{\algofull}{Fast Adaptation through Meta-Learning Embeddings} 

\title{\LARGE \bf Fast Online Adaptation in Robotics through Meta-Learning \\ Embeddings of Simulated Priors}

\author{Rituraj Kaushik, Timoth\'ee Anne and Jean-Baptiste Mouret$^*$
\thanks{*Corresponding author: {\tt\small jean-baptiste.mouret@inria.fr}}
\thanks{All authors have the following affiliations:}
\thanks{Inria, CNRS, Universit\'e de Lorraine, Nancy, France}
\thanks{This work received funding from the European Research Council (ERC) under the European Union’s Horizon 2020 research and innovation programme (GA no. 637972, project ``ResiBots''),  the Lifelong Learning Machines program (L2M) from DARPA/MTO under Contract No. FA8750-18-C-0103, and the Direction G\'en\'erale de l'Armement (project ``Humano\"ide R\'esilient'').}
\thanks{Video: \url{http://tiny.cc/famle_video}}
}

\begin{document}
\AddToShipoutPicture*{\put(50,740){\parbox[b][\paperheight]{\paperwidth}{%
\vfill
\footnotesize
\textbf{2020 IEEE/RSJ International Conference on Intelligent Robots and Systems (IROS),\\ October 25-29, 2020, Las Vegas, NV, USA (Virtual)} 
}}}
\maketitle
\IEEEpeerreviewmaketitle

\begin{abstract}

Meta-learning algorithms can accelerate the model-based reinforcement learning (MBRL) algorithms by finding an initial set of parameters for the dynamical model such that the model can be trained to match the actual dynamics of the system with only a few data-points. However, in the real world, a robot might encounter any situation starting from motor failures to finding itself in a rocky terrain where the dynamics of the robot can be significantly different from one another. In this paper, first, we show that when meta-training situations (the prior situations) have such diverse dynamics, using a single set of meta-trained parameters as a starting point still requires a large number of observations from the real system to learn a useful model of the dynamics. Second, we propose an algorithm called \algo{} that mitigates this limitation by meta-training several initial starting points (i.e., initial parameters) for training the model and allows the robot to select the most suitable starting point to adapt the model to the current situation with only a few gradient steps. We compare \algo{} to MBRL, MBRL with a meta-trained model with MAML, and model-free policy search algorithm PPO for various simulated and real robotic tasks, and show that \algo{} allows the robots to adapt to novel damages in significantly fewer time-steps than the baselines.

\end{abstract}

\section{Introduction}
\label{sec:intro}
Reinforcement learning (RL) algorithms have shown many promising results, starting from playing Atari games from observing pixels or defeating professional Go players. However, these impressive successes were possible due to enormous interaction time with the games in simulation. For instance, in~\cite{heess2017emergence}, around 100 hours of simulation time (more if real-time) was required to train a 9-DOF mannequin to walk in the simulation. Similarly, 38 days of real-time game-play was required for Atari 2600 games~ \cite{mnih_human-level_2015}. The data-hungry nature of these algorithms makes them unsuitable for learning and adaptation in robotics, where the data is much more scarce due to the slow nature of the real physical systems compared to simulated environments. Scarcity of data makes it even more challenging when a robot has to adapt online during its mission because of sudden changes in the dynamics due to component failure (e.g., damages joints), environmental changes (e.g., changes in terrain conditions) or external perturbations (e.g., wind), etc. We refer to these events as different \emph{situations} of the robot.

\begin{figure}[t]
   \centering
   \includegraphics[width=1.0\linewidth]{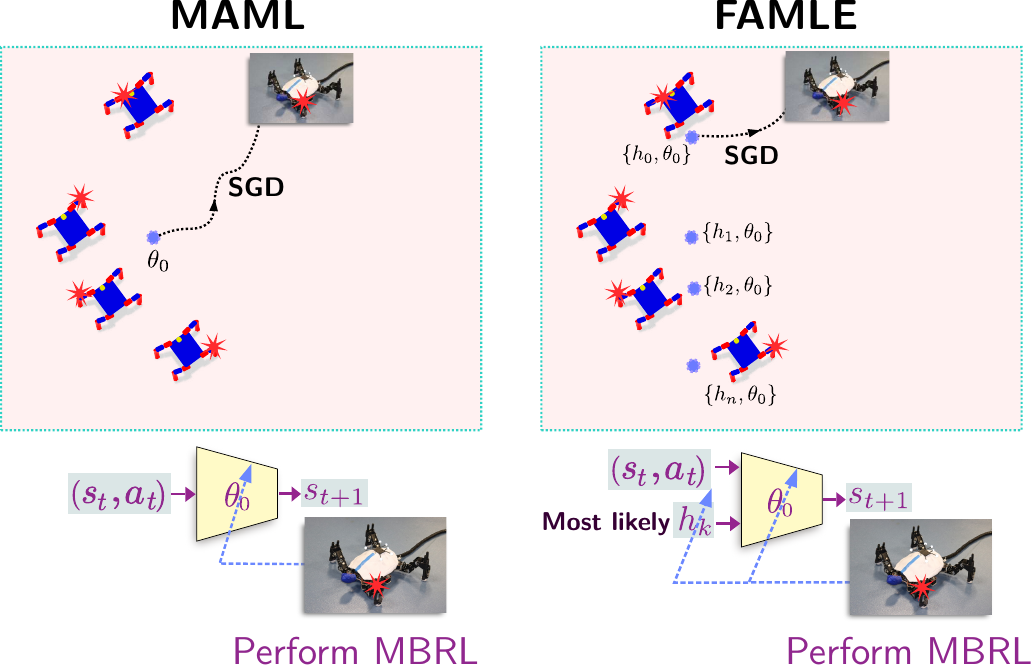}
   \vspace{-1.5em}
   \caption{\label{fig:basic_overview} Basic overview: Compared to model-agnostic meta-learning (MAML) \cite{finn2017model}, \algo{} meta-learns several initializations for the dynamical model corresponding to various situations (e.g., damage conditions) in the simulation. In \algo{}, this is achieved by using a situation conditioned dynamical model, for which, both the initial situation embeddings ($h_{i=0:n}$), as well as the initial model parameters ($\theta_0$), are jointly meta-trained in such a way that the model can be adapted to similar situations with only a few gradient steps. For model-based RL (MBRL) on the real robot, \algo{} figures out the most suitable embedding out of all the trained embeddings to adapt the model using the real world data.}  
   \vspace{-1.5em}
 \end{figure}  

When data-efficiency is crucial for learning and it is possible to learn a useful dynamical model of the robot from the data, then model-based RL (MBRL) algorithms can be a promising direction \cite{chatzilygeroudis2018survey}. The MBRL algorithms iteratively learn a dynamical model of the robot from the past observations, and using that model as a surrogate of the real robot they either optimize the policy \cite{deisenroth2011pilco, chatzilygeroudis2017black, kaushik2018multi} or a sequence of future actions (as in model predictive control) \cite{williams2017information,nagabandi2018learning,chua2018deep}. Since MBRL algorithms draw samples from the model instead of the real robot during the policy optimization, these algorithms can be highly data-efficient compared to model-free RL algorithms. 

Nevertheless, the data requirement of MBRL algorithms typically scales exponentially with the dimensionality of the input state-action space \cite{keogh2010curse,chatzilygeroudis2018survey}. As a consequence, for a relatively complex robot, a typical MBRL algorithm still requires a prohibitive interaction time (from several hours to days) to collect enough data to learn a model of dynamics that is good enough for policy optimization \cite{chua2018deep, nagabandi2018neural}. For example, using MBRL approach, an 8-DoF simulated ``ant'' (actually a quadruped) required around 30 hours of real-time interaction to learn to walk \cite{nagabandi2018neural}. By contrast, we expect robots to adapt in seconds or, at worst, in minutes when they need to adapt to a new situation \cite{cully_robots_2015,chatzilygeroudis2018survey}.


\begin{figure*}[h!]
   \centering
   \includegraphics[width=1.0\linewidth]{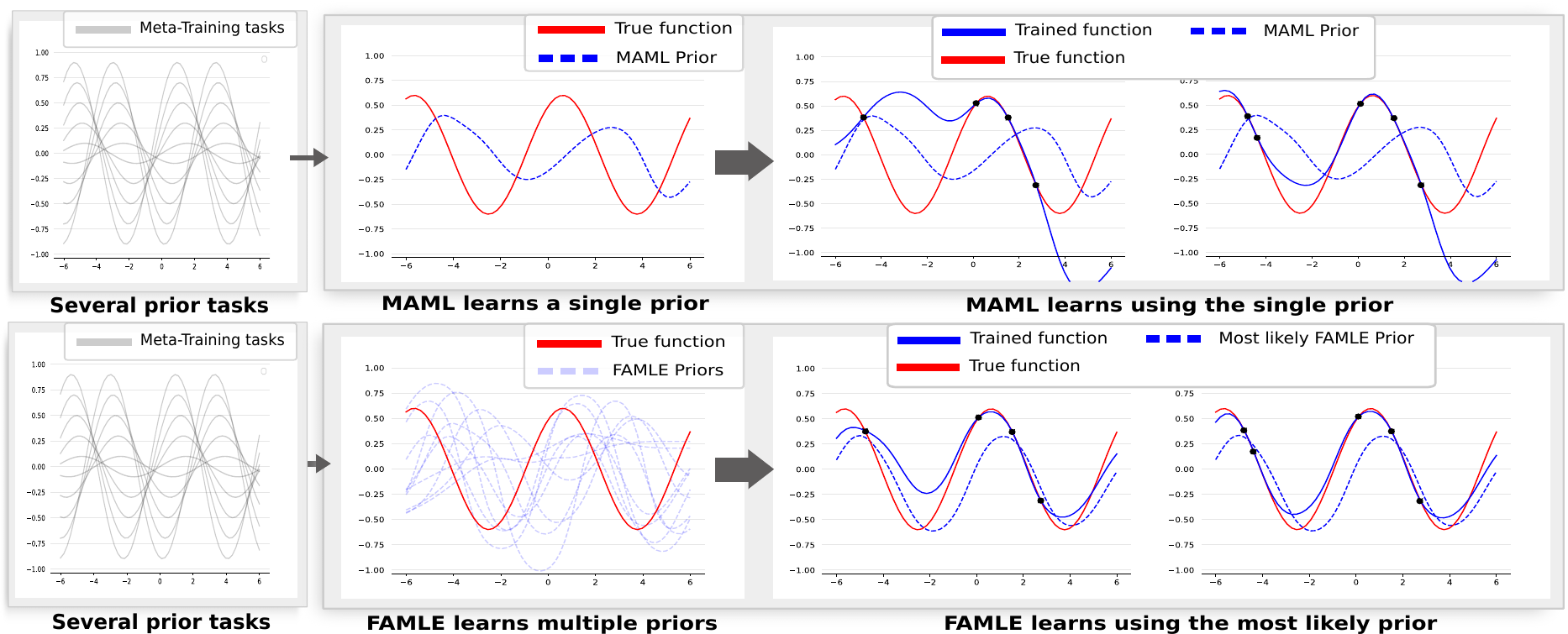}
   \caption{\label{fig:famle_vs_maml} \textbf{Using \algo{} and MAML to learn a simple 1-dimensional sine wave.} Learning several meta-learned priors for the model and selecting the most suitable one for the given data (FAMLE approach) improves data-efficiency. The figure shows that using multiple priors, \algo{} could fit a better function (i.e., closer to the actual function) with 4 and 5 data points than using single priors learned using MAML.}  
   \vspace{-1.5em}
 \end{figure*}  

Unlike robots, animals adapt to sudden changes (e.g., broken bones, walking for the first time on a snowy terrain) almost immediately. Such a rapid adaptation is possible because animals never learn from scratch; instead, they use their past experience as priors or biases to learn faster. Taking inspiration by this phenomenon, meta-learning approaches use past experiences of the robot in various situations to allow it to learn a new but similar skill or to adapt to a similar situation with only a few observations  \cite{finn2017model, nagabandi2018learning}. When used with MBRL, a meta-learning algorithm such as MAML~\cite{finn2017model} finds the initial parameters for the dynamical model from where the model can be adapted to similar situations by taking only a few gradient steps. In other words, meta-learning exploits the similarity among the various past experiences to find out a single prior on the initial parameters.

When a robot has to adapt to many different situations (from broken limb to novel terrain conditions), the prior situations used to meta-train the dynamical model can be diverse and without any substantial global similarity among themselves. For example, a 6-legged robot with a broken leg might experience a very different dynamics than the same intact robot walking on rocky terrain. In fact, we observe that when the prior situations are diverse and do not possess a strong global similarity among themselves, using meta-learning to find a single set of initial parameters for the dynamical model is often not enough to learn quickly. 

One solution to this problem is to find several initial starting points (i.e., initial model parameters) that are meta-trained in such a way that when the model is initialized with the suitable one, the model can be adapted to the real situation of the robot by performing only a few gradient steps using the past observations. However, the question that arises here is how to meta-train several sets of initial parameters, while still generalizing to situations that were not in the training set. In this work, we propose to achieve this objective by using a single dynamical model that is shared among all the prior situations, but takes an additional input that is learned so that it corresponds to the situations, which makes it a situation conditioned dynamical model (Fig. \ref{fig:basic_overview} and \ref{fig:famle_vs_maml}). 


To be more precise, our conditional dynamical model not only takes the current-state and action as inputs, but also a $d$-dimensional vector. This vector is called an embedding of the situation or simply \emph{situation-embedding}. We consider the (initially unknown and randomly set) situation-embeddings as situation-specific parameters of the model, and jointly meta-train all these embeddings as well as the shared model-parameters. In effect, we obtain several meta-trained starting points for the model adaptation -- one for each prior training situation (Fig. \ref{fig:famle_vs_maml}). On the real robot, first, we initialize the model with the meta-trained parameters, then we select the most suitable meta-trained embedding. With the selected embedding as input, we jointly update the embedding as well as the model parameters using gradient descent according to the recent data from the robot. It is to be noted that such joint training of embeddings and model parameters is widely used to learn word embeddings in natural language processing~\cite{collobert2011natural}.

In summary, our main contribution is an algorithm called \algo{} (\algofull) that combines meta-learning and situation embeddings to be able to adapt a dynamical model quickly to a new situation. This model can then be used for model-based RL to optimize its future actions for a given task.
\algo{} can be summarized in two steps (Fig. \ref{fig:concept}):  
\begin{figure*}[h!]
   \centering
   \includegraphics[width=1.0\linewidth]{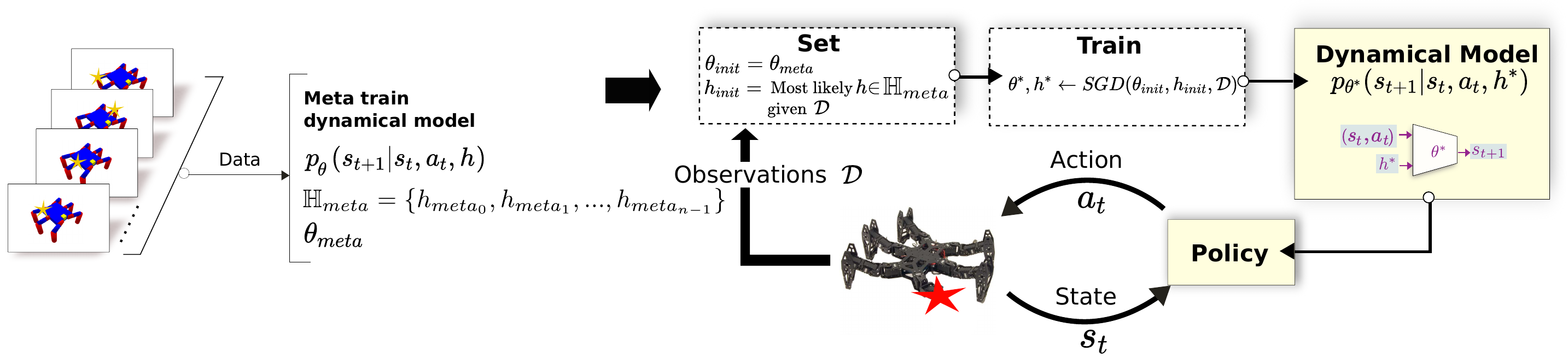}
   \vspace{-1.0em}
   \caption{\label{fig:concept} Overview of \algo{}: first, state-transition data is collected from simulations for $n$ different situations of the robot. Then, the model (i.e., the dynamical model) parameters $\theta$ and situation embeddings $\mathbb{H}=\{h_0,..., h_{n-1}\}$ are meta-trained on the simulated data. On the real robot, when the data is available, \algo{} uses the most suitable embedding from the set $\mathbb{H}$ to adapt the model to the current situation. \algo{} iteratively updates the meta-trained model parameters jointly with the selected embedding using new data and utilizes this model for model predictive control policy.}
   \vspace{-1em}
\end{figure*}  

\begin{itemize}
   \item \textbf{Meta-training:} Generate simulated data for $N$ different situations of the robot, e.g., with a broken joint, on rough terrain, and so on. Meta-train the model-parameters jointly with the embeddings for each of the simulated situations to have $N$ meta-trained embeddings and one set of meta-train initial model parameters. 
   \item \textbf{Online adaptation:} On the real robot, at every $K$ step, initialize the model with the meta-trained parameters and the most likely situation embedding out of the $N$ meta-trained embeddings based on the past $M$ observations. Then update the model jointly with the embedding with gradient descent using the past $M$ observations. This model is then used with a model-predictive-control (MPC) policy for $K$ steps. The process repeats until the task is solved. Please note here that the situation on the real robot is not known a priori and therefore that situation is unlikely to have been exactly simulated in the meta-training phase; this is why the model must be adapted by gradient descent with real data.
\end{itemize}


We compare \algo{} with (a) MBRL with a neural-network dynamics model which was pre-trained with model-agnostic meta-learning (MAML) (b) MBRL with neural-network dynamics model learned from scratch~\cite{nagabandi2018neural}, and (c) proximal policy optimization (PPO)~\cite{schulman2017proximal} on two simulated robots and one real physical quadruped robot. We show that \algo{} allows the robots to adapt to novel damages in significantly fewer time-steps than the baselines. Additionally, we demonstrate that using \algo{}, a physical Minitaur robot (an 8-DoF quadruped) can learn to walk in a minute of interaction in the real world.

\section{Related Work}
\subsection{Model-based Learning in Robotics}
Model-based RL (MBRL) algorithms are some of the most data-efficient learning algorithms in robotics \cite{chatzilygeroudis2018survey}. The core idea of MBRL is to iteratively learn a model of the dynamics of the system and use that model either to optimize a policy (called model-based policy search, MBPS) or to optimize the future sequence of action directly (called adaptive model-predictive control, AMPC). In the MBPS framework, many algorithms such as PILCO~\cite{deisenroth2011pilco}, Black-DROPS~\cite{chatzilygeroudis2017black}, Multi-Dex~\cite{kaushik2018multi} have shown promising results towards data-efficient learning in robotics using Gaussian Processes dynamics model. On the other hand, in the AMPC framework,~\cite{chua2018deep} and \cite{nagabandi2018neural} used deep neural networks to learn the dynamical model. Unlike MBPS, which requires a relatively accurate model of the system dynamics to optimize the policy, AMPC, on the other hand, can use a less precise model of the dynamics as it optimizes the action at every control-step. In order to deal with the model inaccuracies due to small number of samples, the majority of the recent state-of-the-art MBPS algorithms consider model uncertainties for policy optimization~\cite{deisenroth2011pilco, chatzilygeroudis2017black, kaushik2018multi}. Nevertheless, for relatively complex robots, these algorithms still require prohibitively long interaction time (from several hours to days) with the real system when it comes to adapt online to a new situation that perturbs its dynamics, such as damage to joints or a new terrain condition.  

\subsection{Using Priors for Data-efficient Learning}
In order to accelerate the learning process, many recent work leverage prior knowledge about the system dynamics. In traditional robotics, data efficiency is achieved by simply identifying the tunable parameters of a mathematical model of the system using the observed data from the real robot \cite{Hollerbach2016}. In more recent approaches, a parametric or fixed model is ``corrected'' with a non-parametric model (e.g., Gaussian Processes model) to capture potentially non-linear effects in the dynamics of the system. In particular, model-based policy search algorithm like PILCO \cite{deisenroth2011pilco} or Black-DROPS \cite{chatzilygeroudis2017black} can be combined with simulated priors and learn to control a cart-pole in 2 to 5 trials \cite{cutler_efficient_2015, Saveriano2017DataefficientCP,chatzilygeroudis2018using}. These approaches learn a ``residual model'' of the dynamics of the system with Gaussian Processes (GP), i.e., the difference between the simulated and real robot instead of learning the system dynamics from scratch. However, the GP based approaches are limited by the scalability issue of the GP models. Moreover, it is difficult to parameterize different situations that the robot might face in the real world, such as broken legs or faulty actuators, and so on.

Another approach that incorporates priors coming from the simulator to accelerate the learning process is repertoire-based learning. Their key principle is to first learn a large and diverse set of policies in simulation with a ``quality diversity'' algorithm \cite{mouret_illuminating_2015,pugh2016quality,cully2018quality}. Then use an optimization or search process to pick the policies that works best in the current situation \cite{cully_evolving_2015,chatzilygeroudis2018reset}. Repertoire-based learning algorithm such as APROL \cite{kaushik2019adaptive} and RTE \cite{chatzilygeroudis2018reset} learn how the outcomes of these policies change in the real world compared to the simulated world using GP model, where they use the simulated outcomes as prior mean function to the GP model. In particular, APROL uses several repertoires generated for various situations in the simulation. During the mission, APROL tries to estimate the most suitable repertoire to learn the policy outcomes for the real robot. APROL has been able to show promising results in fast online adaptation where a damaged hexapod robot with 18 joints could learn to reach a goal location in less than two minutes of interaction despite the damage as well as the reality gap between the simulation and the real world. Approaches like APROL and MLEI \cite{pautrat2018bayesian} show that using the \emph{right prior} out of many priors could significantly improve the data-efficiency in robot learning. However, the repertoire-based learning approaches require expert knowledge to specify the outcome space. Additionally, it is often not possible to learn the dynamics in the outcome space when the outcome strongly depends on the full-state of the system (e.g., a robotic arm where the outcome space is the end-effector position). 

Recently, gradient-based meta-learning approaches such as MAML \cite{finn2017model} showed a promising direction towards data-efficient learning in robotics using deep neural networks. MAML optimizes the initial parameters for a differentiable model (e.g., a neural-network) of the dynamics such that the model can be adapted to match the actual model of the robot by taking only a few gradient descent steps. In particular, \cite{nagabandi2018learning} applied MAML to online learning scenarios using model-predictive control with the learned dynamical model. Many recent works on MBRL showed that using a hierarchical dynamical model conditioned on the latent variable gives superior data-efficiency \cite{perez2018efficient, saemundsson2018meta}. These approaches pre-train the dynamical model and the prior distribution over the latent variable for the data gathered from various scenarios. Contrary to these work, we cast our online adaptation problem into ``learning-to-learn'' framework, similar to MAML \cite{finn2017model}, where the goal is to learn the initial model-parameters and the initial situation-embeddings using gradient descent in such a way that the model as well as the embeddings can be adapted to unseen situations easily. In effect, our approach produces several initial starting points for the model, each of which can be thought of as a unique \emph{prior} for future adaptation. Selecting the most suitable prior based on the observed data allows us to adapt the model quickly to the real situation.



\section{Preliminaries}
\subsection{Model-Based Reinforcement Learning}
The aim of a RL agent is to maximize the cumulative reward by performing actions in the environment. Formally, a RL problem is represented by a Markov decision process (MDP) which is defined by the tuple $(\mathcal{S}, \mathcal{A}, p, r, \gamma, \rho_0, H)$; where, $\mathcal{S}$ is the set of states, $\mathcal{A}$ is the set of actions, $p(s'|s,a)$ is the state transition probability for given state $s$ and action $a$, $r:\mathcal{S} \times \mathcal{A} \mapsto \mathbb{R}$ is the reward function, $\rho_0$ is the initial state distribution, $\gamma$ is the discount factor, and $H$ is the horizon. An RL agent tries to find a policy $\pi:\mathcal{S} \mapsto \mathcal{A}$ to maximize the expected return $R$ given by:
\begin{align}       
   R &= \mathbb{E}_\pi \Big[\sum_{t=0}^{H-1} r(s,a)\Big] \label{eg:expectedReturn}
\end{align}       
In model-based RL, the above problem is solved by learning the state transition probability function $p(s'|s,a)$ using a function approximator $p_{\theta}(s'|s,a)$, which is also called the dynamical model of the system. The model parameter $\theta$ is optimized to maximize the log-likelihood of the observed data $\mathcal{D}$ from the environment. This model is then used either to optimize a sequence of actions (as in model-predictive control) or to optimize a policy so that the equation \ref{eg:expectedReturn} can be maximized.

\subsection{Gradient-based meta-learning}
Meta-learning approaches assume that the previous meta-training tasks and the new tasks are drawn from the same task distribution $p(\mathcal{T})$, and these tasks share a common structure or similarity which can be exploited for fast learning. Gradient-based meta-learning such as model-agnostic meta-learning (MAML) \cite{finn2017model} tries to find the initial parameters $\theta$ for a differentiable parametric model so that taking only a few gradient descent steps from the initial parameters $\theta$ produces an effective generalization to the new learning task. More concretely, MAML tries to find an initial set of parameters $\theta$ such that for any task $\mathcal{T} \sim p(\mathcal{T})$ with corresponding loss function $\mathcal{L}_\mathcal{T}$, the learner has a low loss after $k$ updates:
\begin{align}
   \min_\theta \mathbb{E}_\mathcal{T} \Big[ \mathcal{L}_\mathcal{T}(U_\mathcal{T}^k(\mathbf{\theta})) \Big] 
\end{align}
where $U_\mathcal{T}^k(\mathbf{\theta})$ is the the update rule (e.g., gradient descent) that updates the parameters $\mathbf{\theta}$ for $k$ times using the data sampled from $\mathcal{T}$. MAML optimizes this problem with stochastic gradient descent as:
\begin{align}
   \mathbf{\theta} &:= \mathbf{\theta} - \alpha \nabla_\mathbf{\theta} \mathcal{L}_\mathcal{T}(U_\mathcal{T}^k(\mathbf{\theta})) \\
   &:= \mathbf{\theta} - \alpha \nabla_\mathbf{\tilde\theta} \mathcal{L}_\mathcal{T}(\mathbf{\tilde\theta}) \nabla_{\mathbf{\theta}}U_\mathcal{T}^k(\mathbf{\theta}) \text{,~~where } \mathbf{\tilde\theta}=U_\mathcal{T}^k(\mathbf{\theta}) \label{eq:maml_update}
\end{align}
To ease the computation of equation \ref{eq:maml_update}, first-order MAML approximates $U_\mathcal{T}^k(\mathbf{\theta})$ as a constant update of $\mathbf{\theta}$ as $U_\mathcal{T}^k(\mathbf{\theta}) = \mathbf{\theta} + \mathbf{b}$ (where $\mathbf{b}$ is a constant). This approximation simplifies equation $\ref{eq:maml_update}$ as:
\begin{align}
   \mathbf{\theta} &:= \mathbf{\theta} - \alpha \nabla_\mathbf{\tilde\theta} \mathcal{L}_\mathcal{T}(\mathbf{\tilde\theta}) \text{,~~where } \mathbf{\tilde\theta}=U_\mathcal{T}^k(\mathbf{\theta})=\mathbf{\theta} + \mathbf{b} \label{eq:fomaml_update}
\end{align}
Another first-order meta learning approach called Reptile \cite{nichol2018first} tries to find a solution $\mathbf{\theta}$ that is close (in Euclidean distance) to each task $\mathcal{T}$'s manifold of optimal solutions. To achieve this Reptile treats $U_\mathcal{T}^k(\mathbf{\theta}) - \mathbf{\theta}$ as a gradient which gives the SGD update of $\mathbf{\theta}$ as:
\begin{align}
   \mathbf{\theta} &:= \mathbf{\theta} + \beta (U_\mathcal{T}^k(\mathbf{\theta}) - \mathbf{\theta}) \label{eq:reptile_update}
\end{align}
Unlike MAML, Reptile does not require to split the data into training-set and test-set for meta-learning. In this paper, we use Reptile (Eq. \ref{eq:reptile_update}) as our meta optimization algorithm due to its computational efficiency and the ease of implementation.

\section{Approach}
\algo{} involves two steps: (1) meta-learning the situation embeddings and the dynamical model from the data gathered from simulation, and (2) adapting the model as well as the situation-embedding on the real robot for an unseen situation during the mission. In the following subsections, we elaborate these two steps.  

\subsection{Meta-learning the situation-embeddings and the dynamical model}
We consider a predictive model $p_\theta(s_{t+1}|s_t, a_t, h)$ of the dynamics, where $s_{t+1},s_t,a_t$ and $h$ are the current-state, the next-state, the applied action and the situation-embedding corresponding to the current situation of the robot. This is represented by a neural network $f_\theta(s_t,a_t,h)$ that predicts the mean of the distribution of next state. In the real world, the robot might face any situation $c$ which comes from a distribution of situations $p(c)$. Since, in this work, we consider a \emph{situation} as any circumstance that perturbs the dynamics of the system, so $c$ represents any unknown dynamics of the system sampled from the distribution of dynamics $p(c)$. To collect the state-transition data from simulation we perform the following steps:     

Create empty sets $\mathbb{C}$ and $\mathbb{D}$. Now, for $i=1 \text{ to } N$
\begin{enumerate}
   \item Sample a situation $c_i \sim p(c)$ and insert it in the set $\mathbb{C}$, i.e.,  $\mathbb{C} = \mathbb{C} \cup \{c_i\}$
   \item Instantiate a simulator of the robot for the situation $c_i$.
   \item Perform $n$ random actions on the simulated robot and create data-set $\mathcal{D}_{c_i} = \{(s_t, a_t, s_{t+1}) | t=1,\ldots,n\}$  
   \item Save the data-set into $\mathbb{D}$, i.e., $\mathbb{D} = \mathbb{D} \cup \{\mathcal{D}_{c_i}\}$
\end{enumerate}
Then, corresponding to each sampled situation $c_{i=1:N}$, we randomly initialize situation-embeddings $\mathbb{H} = \{h_{c_i}|i=1,\ldots,N\}$. Also, we randomly initialize the model parameter $\theta$. Then the negative log-likelihood loss for any situation $c_i \in \mathbb{C}$ can be written as:
\begin{align}
\mathcal{L}_{\mathcal{D}_{c_i}}(\theta, h_{c_{i=1:N}}) = \mathbb{E}_{\mathcal{D}_{c_i}} \big[- \log p_{\theta}(s_{t+1}|s_t, a_t, h_{c_i})\big] \label{eq:model_loss}
\end{align}  
Our meta-learning objective is to find initial model-parameters $\theta_{meta}$ and situation-embeddings $\mathbb{H}_{meta}$, such that for any situation $c_i \in \mathbb{C}$, performing $k$ gradient descent steps from $\theta_{meta}$ and $\mathbb{H}_{meta}$ minimizes the loss given by equation \ref{eq:model_loss}. This objective can be written as a meta-optimization problem as:
\begin{align}
\theta_{meta}, \mathbb{H}_{meta} = \arg\min_{\theta, h_{c_{i=1:N}}} \mathbb{E}_{c \sim \mathbb{C}} \Big[ \mathcal{L}_{\mathcal{D}_{c}} \big(U_{c}^k(\theta, h_{c}) \big) \Big] \label{eq:meta_objective}
\end{align}
%
where, $U_c^k(\cdot, \cdot)$ is the gradient descent update rule (applied for $k$ gradient descent steps) that updates the parameters $\theta$ and the situation-embedding $h_c$ for any situation $c \in \mathbb{C}$. We optimize the above problem using meta-learning approach similar to Reptile \cite{nichol2018first} (as in Eq. \ref{eq:reptile_update}). However, unlike Reptile update in equation \ref{eq:reptile_update}, we update both $\theta$ and embedding $h_{c_i}$ simultaneously. More precisely, at each update step, we randomly choose a situation $c_i$ from the set of situations $\mathbb{C}$ and perform the following update on $\theta$ and embedding $h_{c_i}$:
\begin{align}
\tilde \theta, \tilde h_{c_i} & = U_{c_i}^k(\theta, h_{c_i}) \\
\theta &:= \theta + \alpha_{meta} (\tilde \theta - \theta) \\
h_{c_i} &:= h_{c_i} + \beta_{meta} (\tilde h_{c_i} - h_{c_i}) \label{eq:famle_update}
\end{align}
where, $\alpha_{meta}$ and $\beta_{meta}$ are the meta-learning rate. At convergence, we obtain the meta-trained parameters $\theta_{meta}$ and the set of situation-embeddings $\mathbb{H}_{meta}$ for each situation in the set $\mathbb{C}$. Combination of these $N$ situation-embeddings and the meta-trained model-parameters will serve as $N$ different priors for future adaption of the dynamical model to unseen situations.
%
\begin{algorithm}[H]
   \scriptsize
   \caption{\algo{}: Meta-learning}
   \label{algo:metalearning}
   \begin{algorithmic}[1]
      \Require $\mathbb{D}=\{\}$ \Comment{Empty set of data-sets}
      \Require $U^k_c(\cdot, \cdot)$ \Comment{$k$ steps SGD update rule for situation $c$}
      \For {$i = 1,2,..., N$}
         \State $c_i \sim p(c)$  \Comment{Sample a situation}
         \State $\mathbb{C} \leftarrow c_i$ \Comment{Save the situation}
         \State  $\mathcal{D}_{c_i} = \{(s_t, a_t, s_{t+1}) | t=1,\ldots,n\}$ \Comment{Simulate and collect data}
         \State $\mathbb{D} = \mathbb{D} \cup \{D_{c_i}\}$ \Comment{Save the data-set for situation $c$} 
      \EndFor
      \State Randomly Initialize $\theta$ \Comment{Model parameters}
      \State Randomly Initialize $\mathbb{H}=\{h_{c_i} | i=1,\ldots, N\}$ \Comment{Situation embeddings}
      \For {$m = 0,1,...$}
         \State ~$\mathcal{D}_{c_i} \sim \mathbb{D}$ \Comment{Sample a data-set}
         \State  $\tilde \theta, \tilde h_{c_i} = U_{c_i}^k(\theta, h_{c_i})$ \Comment{Perform SGD for k steps}
         \State  $\theta := \theta + \alpha_{meta} (\tilde \theta - \theta)$ \Comment{Move $\theta$ towards $\tilde \theta$}
         \State  $h_{c_i} := h_{c_i} + \beta_{meta} (\tilde h_{c_i} - h_{c_i})$ \Comment{Move $h_{c_i}$ towards $\tilde h_{c_i}$}
      \EndFor
      \State Return $\theta, \mathbb{H}$ \Comment{Return meta-trained parameters and embeddings}   
   \end{algorithmic}
\end{algorithm}

\subsection{Online adaptation to unseen situation}
As the robot might face any situation that can perturb its dynamics during the mission, we want to learn a new dynamical model after every $K$ control steps using $M$ recent observations. To learn this model, we set the meta-trained parameters $\theta_{meta}$ in the model and compute the likelihood of the $M$ recent observations for each situation embedding in $\mathbb{H}_{meta}$. Then, we use the embedding that maximizes the likelihood of the recent data and train the model parameters as well as the selected embedding. To be more precise, if $\mathcal{D}_M$ is the recent $M$ observations on the robot, then:
\begin{align}
h_{Likely} = \arg\max_{h \in \mathbb{H}_{meta}} \mathbb{E}_{\mathcal{D}_M} \big[ \log p_{\theta_{meta}}(s_{t+1}|s_t, a_t, h)\big]
\end{align}
Then we simultaneously update both model parameters $\theta = \theta_{meta}$ and the most likely situation-embedding $h = h_{Likely}$ with by taking $k$ gradient steps:
\begin{align}
\theta &:= \theta - \alpha \nabla_\theta \mathcal{L}_{\mathcal{D}_M} (\theta, h) \nonumber\\
h_c &:= h_c - \beta \nabla_{h_c} \mathcal{L}_{\mathcal{D}_M} (\theta, h) \label{eq:famle_update}
\end{align}

After this optimization, we get the optimized model parameters $\theta^*$ and situation embedding $h^*$ yielding the model $p_{\theta^*}(s_{t+1}|s_t,a_t,h^*)$. Now using this model, the model predictive control (MPC) method can be used as a policy to maximize the long term reward. In this work, we consider random-sampling shooting \cite{rao2009survey} to optimize the sequence of action using the model. This method is computationally faster compared to other sampling-based methods such as CEM \cite{botev2013cross} and relatively easy to implement as well as parallelize. Additionally, due to the randomness, it allows implicit exploration in state-action space, which helps to learn a better model. Random-sampling shooting has been successfully demonstrated as an action sequence optimization method for MPC in recent robot learning papers such as~ \cite{nagabandi2018neural}. At any state $s$, the next action for the robot is optimized as follows:
\begin{enumerate}
\item Sample $N$ random trajectories of action where each action is sampled from a uniform distribution: $\{\tau_i | i=0,..., N-1\}$ and $\tau_i=(a^i_0, a^i_1, ..., a^i_{H-1})$
\item Evaluate the trajectories on the model $f_{\theta^*}(s,a,h^*)$ and select the one that maximizes the total reward.
\begin{align}
   &\tau^* = \arg\max_{\tau_i} \sum_{t=0}^{H-1} r(s_t, a_t, f_{\theta^*}(s_t,a_t,h^*)) \\ 
   &\text{Where, $r(\cdot, \cdot, \cdot)$ is the reward function.} \nonumber
\end{align}
\item Apply the first action of $\tau^*$ and repeat from step (1) until the task is solved.
\end{enumerate}

Pseudo codes for \algo{} is given in Algorithm \ref{algo:metalearning} and Algorithm \ref{algo:adaptation}.


\begin{algorithm}[H]
   \scriptsize
   \caption{\algo{}: Fast adaptation \& control}
   \label{algo:adaptation}
   \begin{algorithmic}[1]
      \Require $\theta_{meta}, \mathbb{H}_{meta}$ \Comment{Meta-learned initial model parameters and embeddings}
      \Require $\mathcal{D}_{M}=\phi$ \Comment{Empty set for $M$ recent observations}
      \Require $r(\cdot, \cdot, \cdot)$ \Comment{Reward function}
      \While {not $Solved$}
         \State $h_{Likely}$ = most likely $h_{c_i} \in \mathbb{H}_{meta}$ given $\mathcal{D}_{M}$ and $\theta_{meta}$
         \State $\theta^*, h^* = k \text{ steps SGD from } \theta_{meta}, h_{Likely} \text{ using } \mathcal{D}_{M}$
         \State Apply optimal action $a = MPC(\theta^*, h^*, r(\cdot, \cdot, \cdot))$
         \State $\mathcal{D}_{M} \leftarrow \mathcal{D}_{M} \cup \{(s_t, a_t, s_{t+1})\}$ \Comment{Insert observation}
         \State \textbf{if } $size(\mathcal{D}_{M}) > M$ \textbf{ then } Remove oldest from $\mathcal{D}_{M}$
      \EndWhile
   \end{algorithmic}
\end{algorithm}

\section{Experimental Results}
Here, our goal is to evaluate the data-efficiency of \algo{} and compare it to various baseline algorithms. As a metric for data-efficiency, we focus on the real-world interaction time (or time-steps) required to learn a task by a robot. So, a highly data-efficient algorithm should require fewer time-steps to achieve higher rewards in a reinforcement learning set-up. We compared \algo{} on various tasks against the following baselines and showed that \algo{} requires fewer time-steps to achieve higher rewards than the baselines:
\begin{itemize}
   \item \textbf{PPO: } Proximal Policy Optimization, a model-free policy search algorithm, which is easy to implement, computationally faster, and performs as good as current state-of-the-art model-free policy search algorithms.  
   \item \textbf{AMPC: } Adaptive MPC, i.e., MPC using an iteratively learned dynamical model of the system from scratch using past observations with a neural network model.
   \item \textbf{AMPC-MAML: } Adaptive MPC with a meta-trained neural network dynamical model. Here, the network is meta-trained using MAML for the same situations that are used in \algo{}. At test time, the model is updated using the recent data with meta-trained parameters as initial parameters of the network.  
\end{itemize}

For the experiments with the physical robots, the simulated robots have very similar dimensions, weights and actuators as those of the real robots. However, we did not explicitly fine tune these parameters in the simulator to match the behavior exactly on real robot. For all the MBRL algorithms, we used neural networks that predict the \emph{change in current state} of the robot. To generate the data for the prior situations, we used pybulet physics simulator \cite{coumans2013bullet}. The code\footnote{\scriptsize Code:\url{https://github.com/resibots/kaushik_2020_famle}}and the video\footnote{\scriptsize Video:\url{http://tiny.cc/famle_video}} of the experiments can be found online.
\subsection{Goal reaching with a 5-DoF planar robotic arm}
In this simulated experiment, the end-effector of the velocity-controlled ($10$ Hz) arm has to reach a fixed goal as quickly as possible. The joints of the arm might have various damages/faults: (1) weakened motor (2) wrong voltage polarity, i.e., opposite rotation compared to a normal joint, and (3) dead/blocked motor. Here, the state-space and action space have 12 and 5 dimensions, respectively. The embedding vector size for \algo{} was 5. The dynamical models were learned using neural networks with 2 hidden layers of size 70 and 50. For AMPC-MAML and \algo{}, the models were meta-trained on the simulation data for 11 different damage situations. For testing, two random damages were introduced on the arm, which were not in the meta-training set. Experiments were performed on 30 replicates, sharing the same test damage condition for all the replicates. 
\begin{figure}[h!]
   \centering
   \includegraphics[width=1.0\linewidth]{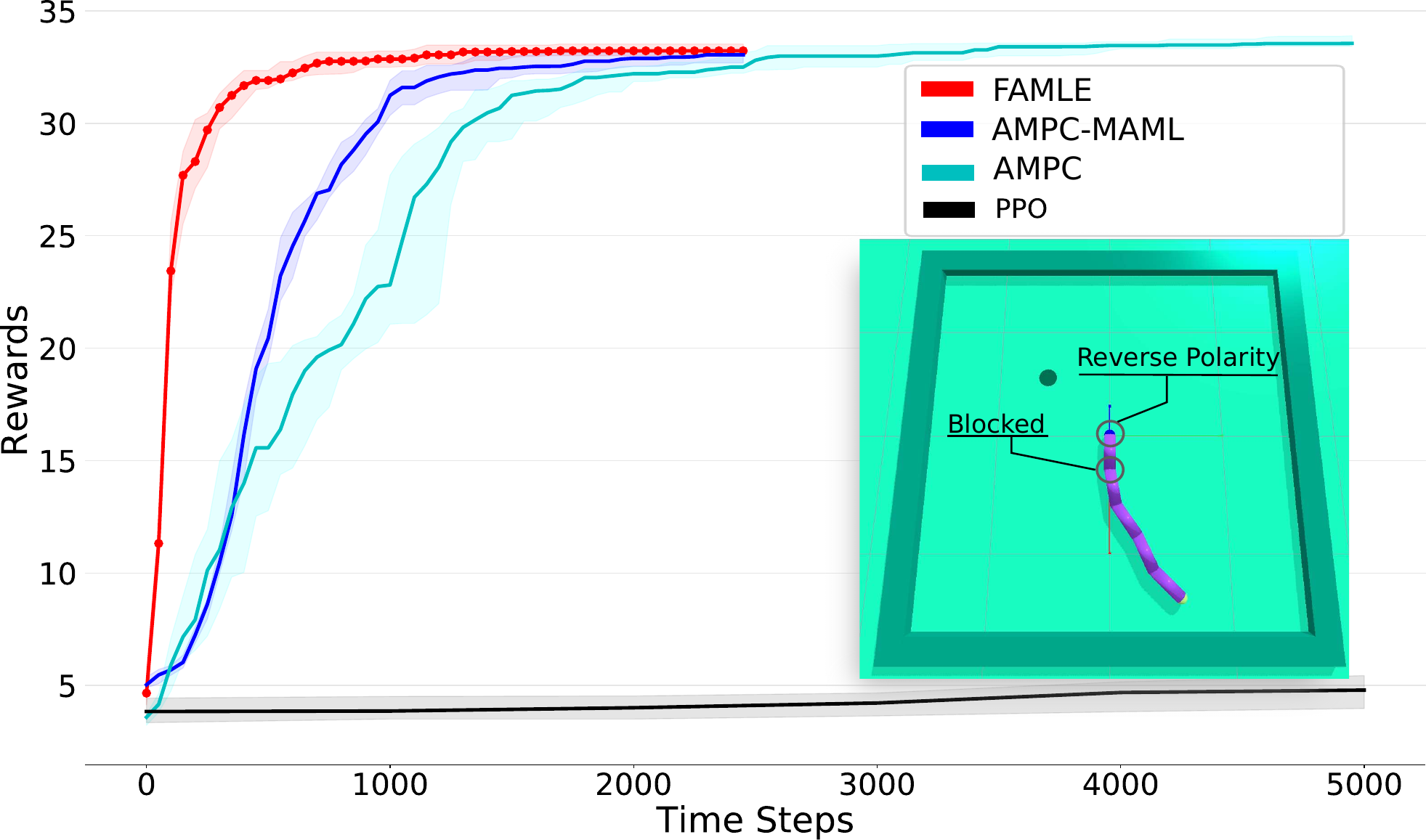}
   \vspace{-1.5em}
   \caption{\label{plot:reacher} Goal reaching task: the 5-DoF planer arm has to reach target in minimum number of steps despite damage in its joints. Plot shows median, 25 and 75 percentile of the accumulated reward per episode for 30 replicates. Here, \algo{} maximizes the reward in fewer time-steps than the baselines.} 
   \vspace{-1em}
 \end{figure}  
 

The plot (Fig. \ref{plot:reacher}) shows that \algo{} achieves higher reward much faster than the baselines and solves the task (i.e., reward more than 30) in $\sim$500 steps (50 seconds). Here, due to the large variations in the dynamics caused by different damage/fault situations in the meta-training data, meta-trained parameters using MAML could not generalize to all the situations. Thus, it required more data to adapt the model to the current situation during test time. In this task, \algo{} achieved $\sim$100\% improvement in time compared to AMPC-MAML. As expected, being model-free, PPO could not reach the performance of the model-based approaches within the maximum time-steps limit.

 \subsection{Ant locomotion task}

In this task, a 4-legged simulated robot (8 joints, torque-controlled, 100 Hz) has to walk in the forward direction as far as possible to maximize the reward. Here, the robot might have (1) blocked joints and/or (2) error in the orientation measurement (i.e., sensor fault). The state-space and the action-space for this problem are 27 and 8 dimensional, respectively. The embedding vector size for \algo{} was 5. The dynamical model is a neural network with 3 hidden layers of size 200, 200 and 100. Meta-training data for AMPC-MAML and \algo{} was collected (applying random actions) from simulations for 20 different damage/fault situations of the robot. At the beginning of the test, a random joint damage and orientation error (not in the training set) were introduced. 

 \begin{figure}[h!]
   \centering
   \includegraphics[width=1.0\linewidth]{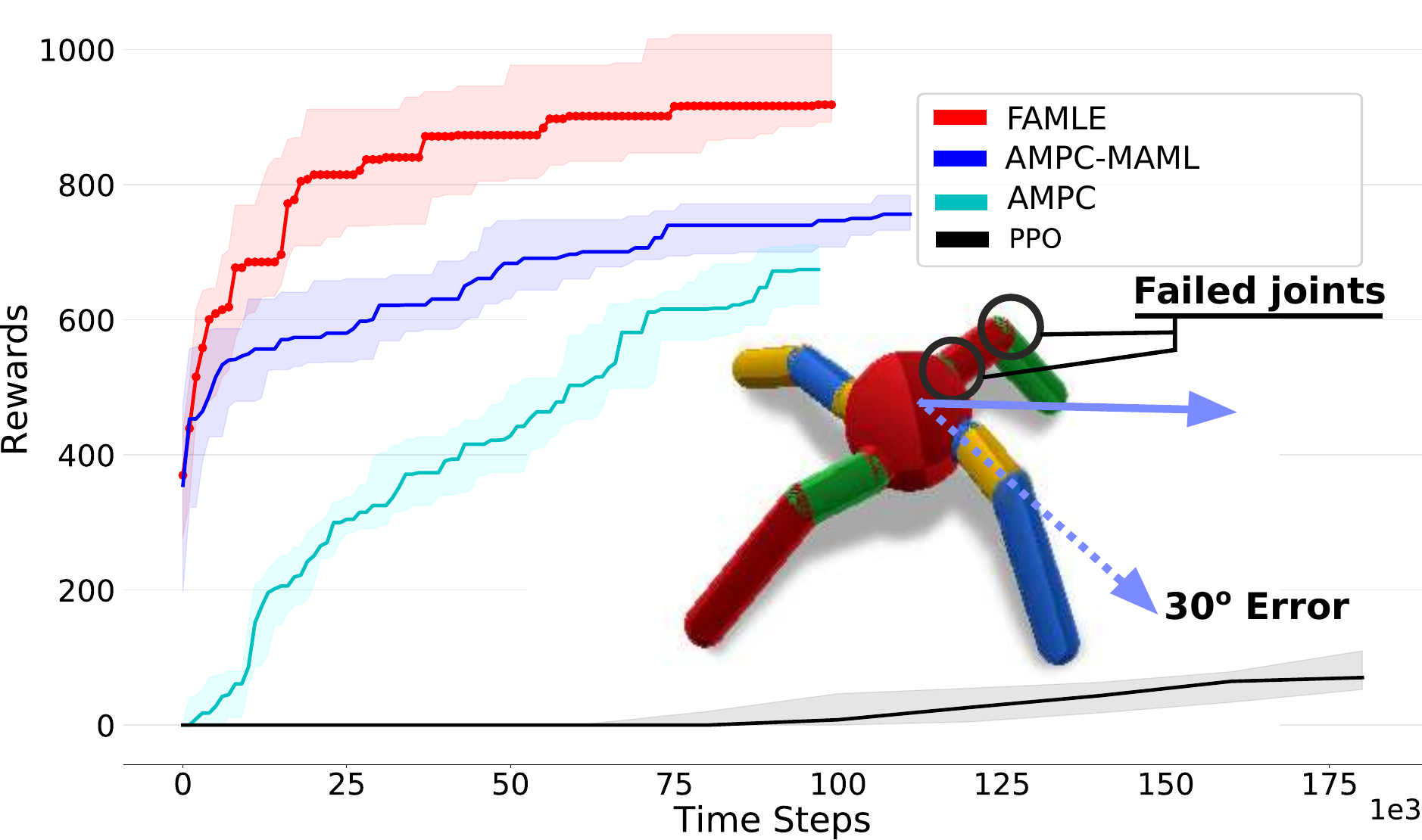}
   \vspace{-1.5em}
   \caption{\label{plot:ant} Ant locomotion task: the damaged ant has to move as far as possible in the forward direction. Plot shows median, 25 and 75 percentile of the accumulated reward per episode for 30 replicates. Here, \algo{} finds higher rewards in much lower time-steps than the baselines.} 
   \vspace{-1em}
 \end{figure}  

 The plot (Fig. \ref{plot:ant}) shows that \algo{} achieves higher reward than the baselines and solves the task (i.e., reward more than 800) in $\sim$25000 steps (4.17 minutes of real-time interaction). Similar to the previous task, AMPC-MAML performed worse compared to \algo{} in this task too. Here, \algo{} achieved $\sim$500\% improvement in time compared to AMPC-MAML. On the other hand, PPO performed the worst.

 \subsection{Quadruped damage recovery}
 In this online adaptation task, a physical quadruped (12 joints) has to recover from damage to its legs and/or faults in the orientation measurement and reach the goal as quickly as possible. Here, the action is a 4-dimensional vector that modulates the gait of the robot though the period functions associated with each leg\footnote{\footnotesize \label{foot:unable_to_learn}In the preliminary experiments, we were unable to learn a full dynamical model of the physical robots that is accurate enough for MPC.}. At every second, a new action is applied that produces a new gait on the robot using a low-level controller. The state of the robot includes the 2D position and 2D orientation (sine and cosine of rotation along the vertical axis). The embedding vector size for \algo{} was 20. The dynamical models were learned using neural networks with 2 hidden layers of size 100. The meta-training data for the dynamical model was collected (by applying random actions) from a low fidelity simulator of the robot for total 20 different situations, each of which includes either a joint block or orientation measurement error between 0 to 360 degrees. We tested 3 different situations on the real physical robot: (1) one blocked leg (2) orientation fault, and (3) one blocked leg as well as orientation fault. Additionally, we also evaluated the performance of \algo{} by introducing orientation fault on the robot online (during the deployment). In all the experiments, the goal was 2.5 meters away from the starting position of the robot.

 \begin{figure}[h!]
   \centering
   \includegraphics[width=1.0\linewidth]{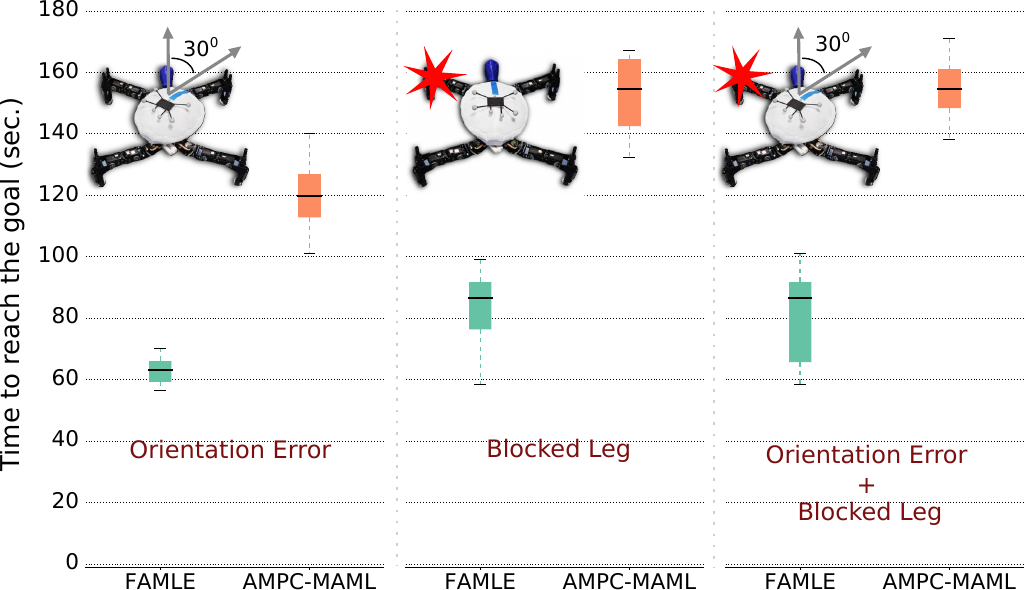}
   \caption{\label{plot:quadruped_plot} Quadruped damage recovery task: Box plot (over 10 replicates) shows time required to reach the goal location. Using \algo{} the robot could adapt to both orientation error as well as leg damage and solve the task in less than 2 minutes of interaction.}  
 \end{figure}  

\begin{table}
   \centering
   \caption{\label{tab:pexodquad} Quadruped damage recovery task}
   \begin{tabular}{ | l | c | c | c |}
   \hline
   \multicolumn{4}{|c|}{ORIENTATION FAULT OF $30^0$} \\
   \hline
    ~~ & Steps & Time (Sec.) & Success Rate \\ 
   \hline
   \algo{} & $32.6 \pm 2.3$ & $63.4 \pm 4.5$ & $100\%$\\ 
   \hline
   AMPC-MAML & $61.8 \pm 7.2$ & $120.1 \pm 13.9 $ & $40\%$\\   
   \hline
   AMPC & - & - & $0\%$ \\
   \hline
   \hline
   \multicolumn{4}{|c|}{ONE BLOCKED LEG} \\  
   \hline
   ~~ & Steps & Time (Sec.) & Success Rate \\ 
   \hline
   \algo{} & $43.1 \pm 6.4$ & $83.8 \pm 12.4$ & $100\%$\\ 
   \hline
   AMPC-MAML & $78.3 \pm 7.2$ & $152.2 \pm 14.0 $ & $40\%$\\   
   \hline
   AMPC & - & - & $0\%$ \\  
   \hline
   \hline
   \multicolumn{4}{|c|}{ONE BLOCKED LEG + ORIENTATION FAULT OF $30^0$} \\  
   \hline
   ~~ & Steps & Time (Sec.) & Success Rate \\ 
   \hline
   \algo{} & $41.6 \pm 7.7$ & $80.9 \pm 15.0$ & $100\%$\\ 
   \hline
   AMPC-MAML & $79.5 \pm 6.1$ & $154.6 \pm 11.9 $ & $40\%$\\   
   \hline
   AMPC & - & - & $0\%$ \\  
   \hline
\end{tabular}
\vspace{-2em}
\end{table}

The table \ref{tab:pexodquad} and box-plot \ref{plot:quadruped_plot} show the comparison of \algo{} with AMPC and AMPC-MAML baselines on the Quadruped damage recovery task. Results show that using \algo{}, the robot could reach the goal $100\%$ of the time by taking significantly less time than the baselines. Using AMPC-MAML, the robot was able to reach the goal only $40\%$ of the time within the maximum allotted time-steps of 80. On the other hand, using AMPC, the robot was never able to reach the goal within the maximum allotted time steps. Here \algo{} achieved $\sim$66\% improvement in time compared to AMPC-MAML.

\subsection{Minitaur learning to walk}
In this experiment, we used the quadruped robot Minitaur from Ghost Robotics. The goal here is to learn to walk in the forward direction as far as possible. Here, the state-space is 6-dimensional (center of mass position and orientation) and action is a 4-dimensional vector (applied at each second) that modulates the gait of the robot through periodic functions associated with each leg (see footnote \ref{foot:unable_to_learn}). The embedding vector size for \algo{} was 8. The dynamical models were learned using neural networks with 2 hidden layers of size 20. For meta-training, we collected state transition data from the simulator of the robot in pybullet (with random actions) for 3 different friction conditions (default, 0.5 times and 2 times of the default friction) and 3 different weights of the base of the robot (real weight, 1.5 times and 2 times of the real weight). 

Due to the high reality gap between the simulated robot and the real robot, a dynamical model trained directly on the data collected from the simulator (for the default weight and friction) performs poorly on the real robot. To verify this, we used such a model on the Minitaur in the simulator as well as on the real robot. On the simulated robot, the robot could immediately walk in the forward direction using model predictive control. However, on the real robot, the robot could not move and failed due to exceeding current limits in the motors. Now, to evaluate the meta-trained model using \algo{}, we used it on the real robot. Thanks to the meta-trained embeddings, the robot could quickly figure out the most suitable embedding to update its model from the real observations. With \algo{}, using only 60 data points (1 minute of real interaction) from the real robot, the robot could consistently walk forward without fail (see Table \ref{tab:minitaur}). Here \algo{} achieved $\sim$50\% improvement in time (to reach the maximum distance) compared to AMPC-MAML.


\begin{table}
   \centering
   \caption{\label{tab:minitaur} Minitaur learning to walk}
   \begin{tabular}{ | l | c | c | c |}
   \hline
    ~~ & Max distance & Interaction time \\ 
   \hline
   \algo{} & $4.8$ meters & $60$ seconds \\ 
   \hline
   AMPC-MAML & $1.6$ meters & $90$ seconds \\   
   \hline
   AMPC & $0.3$ meters & $130$ seconds \\
   \hline
\end{tabular}
\vspace{-1.5em}
\end{table}

\section{Conclusion}

\algo{} speeds up model learning by 50\% to 500\% compared to MAML \cite{finn2017model}, which already leads a large improvement compared to learning the model from scratch \cite{nagabandi2018learning}. \algo{} made the most difference in the Ant locomotion task ($\sim$500\% improvement) and the least difference in the Minitaur learning to walk task ($\sim$50\% improvement). The reason is that the state-action space is much larger (35D and 17D respectively) in the Ant and the robotic arm tasks compared to the real robotic tasks (8D and 10D in the Quadruped and Minitaur task respectively), which makes model learning easier in the latter case. Overall, the results demonstrate that \algo{} can leverage the meta-learned embeddings to select the most suitable starting points for model learning, which accelerates learning even more than single-starting point meta-learning (MAML). 

The ability to adapt rapidly to unforeseen situations is one of the main open challenges for robotics. One of the key component to achieve such rapid adaptation is the effective use of prior knowledge \cite{chatzilygeroudis2018survey}: using many prior situations in simulation, \algo{} allows robots to adapt quickly in reality. We believe, \algo{} is a promising direction towards adaptive robots for long-term missions in the real and uncertain world.
\bibliographystyle{IEEEtran}
\bibliography{mybib}

\end{document}